\documentclass[11pt,a4paper]{article}
\usepackage[nohyperref]{naaclhlt2019}
\usepackage{times}
\usepackage{latexsym}

\usepackage{url}
\usepackage{times}
\usepackage{latexsym}
\usepackage{amsmath}
\usepackage{amssymb}
\usepackage{graphicx}
\usepackage{multirow}
\usepackage{subfig}
\usepackage{url}
\usepackage{paralist}
\usepackage{latexsym}
\usepackage{amsmath}
\usepackage{color}
\usepackage{graphicx}
\usepackage{booktabs}
\usepackage{subfig}
\usepackage{url}
\usepackage{tabularx}
\usepackage{multirow}
\usepackage{verbatim}
\usepackage{CJK}
\usepackage{textcomp}
\usepackage{amsmath,bm}
\usepackage{soul}

\newcommand \TODO[1]{{\bf \color{red} \\TODO:#1.\\}}

\makeatother

\usepackage{xcolor}

\makeatletter  
\newif\if@restonecol  
\makeatother

\usepackage[linesnumbered,ruled,vlined]{algorithm2e}
\usepackage{algpseudocode}  
\usepackage{amsmath}  
\SetKwComment{Comment}{$\triangleright$\ }{}
\SetKw{KwInit}{Initialize: }

\aclfinalcopy 

\title{Learning to Discriminate Noises for Incorporating External Information \\ in Neural Machine Translation}

\author{
	Zaixiang Zheng \\ Nanjing University \\ {\normalsize \tt zhengzx@nlp.nju.edu.cn} \\\And
	Shujian Huang \\ Nanjing University  \\ {\normalsize \tt huangsj@nlp.nju.edu.cn} \\\And
	Zewei Sun \\ Nanjing University \\ {\normalsize \tt sunzw@nlp.nju.edu.cn} \\\AND
	Rongxiang Weng \\ Nanjing University \\ {\normalsize \tt wengrx@nlp.nju.edu.cn} \\\And
	Xin-Yu Dai \\ Nanjing University \\   {\normalsize \tt dxy@nlp.nju.edu.cn} \\\And
	Jiajun Chen \\ Nanjing University \\ {\normalsize \tt chenjj@nlp.nju.edu.cn}
}

\date{}

\begin{document}
\maketitle
\begin{abstract}

Previous studies show that incorporating external information could improve the translation quality of Neural Machine Translation (NMT) systems.
However, there are inevitably noises in the external information, severely reducing the benefit that the existing methods could receive from the incorporation.
To tackle the problem, this study pays special attention to the discrimination of the noises during the incorporation.
We argue that there exist two kinds of noise in this external information, i.e. global noise and local noise, which affect the translations for the whole sentence and for some specific words, respectively. 
Accordingly, we propose a general framework that learns to jointly discriminate both the global and local noises, so that the external information could be better leveraged. 
Our model is trained on the dataset derived from the original parallel corpus without any external labeled data or annotation.
Experimental results in various real-world scenarios, language pairs, and neural architectures indicate that discriminating noises contributes to significant improvements in translation quality by being able to better incorporate the external information, even in very noisy conditions.
    
\end{abstract}

\begin{CJK}{UTF8}{gbsn}
    \begin{table}[t]
    \footnotesize
    \renewcommand\arraystretch{1.2}
    \centering
        \begin{tabular}{m{48pt}|m{146pt}} 
            \toprule 
            \textsc{Source}            & 工人(worker)~~正(just)~~在(in)~~搬运(moving)~~设备(equipment)\\ 
            \textsc{Refer.}            & the workers are moving equipment \\
            \textsc{NMT}    & workers are \it{working} on the equipment\\
            \hline 
            \hline
            \textsc{Human}  & {\bf moving}  \\
            \textsc{Dict}  &  worker just in {\bf moving} equipment \\
            \textsc{SMT}  &  workers are {\bf moving} devices   \\
            \hline 
            \hline
            \textsc{Expected}    & workers are moving equipment \\
            \bottomrule
        \end{tabular}
        \caption{The motivating example. The first three lines are the source sentence, the reference translation and the translation from a current NMT system with a translation error (``搬运(moving)" to ``{working}"). The next three lines give examples of three different external information: Human interactive suggestions (\textsc{Human}), word translations generated by a bilingual dictionary (\textsc{Dict}) and the translation from a Statistic Machine Translation system (SMT). Each case contains correct translation that may help the NMT system.
        The last line shows the expected improved translation result of original NMT, in which the wrong translation ``working'' is corrected to ``moving''.}
        \label{table-motivation}
    \end{table}
\end{CJK}

\section{Introduction}
\label{sec:intro}

    Recently, Neural Machine Translation (NMT) systems have achieved state of the art performance in large-scale machine translation tasks~\cite{bahdanau2014neural,sutskever2014sequence,D14-1179,Vaswani2017Attention,Gehring2017ConSeq}. 
    While previous researches mainly aim at designing more sophisticated models to enhance NMT models themselves~\cite{tu-EtAl:2016:P16-1,Mi2016,D16-1027,Weng:2017:EMNLP},
    another way to improve the translation performance is to provide outside assistance to the NMT systems~\cite{chen-EtAl:2017:Long6,Wang2017Cross,wang2017neural,knowles2016neural,zhou-EtAl:2017:Short1}. Here we refer to this outside assistance as the \textit{external information} in general.
    
    The form and content of the external information could be of various kinds, depending on diverse real-world scenarios.
    In Table~\ref{table-motivation}, we show examples of three different kinds of external information. 
    Because the external information could be either long or short, either a whole sentence or several phrases or even just individual words, we propose to use a set of externally given words, called \textit{external words}, as a general form to cover all these kinds of external information. Here external words could be any of the cases in Table~\ref{table-motivation}. 
    
    While previous approaches generally focus on how to integrate external information~\cite{Gu2017,wang2017neural}, less attention is paid to noises in the given information. 
    We argue that neglecting the noises will have adverse impacts on improving the translation quality. 
    Furthermore, we divide the noises into the following two categories: 
    \begin{compactitem}
        \item \textbf{Global Noise}: Words in the external words that are generally irrelevant to the translations of the words of the whole sentence. 
        
        \item \textbf{Local Noise}: At a given translation time step, words in the external words that are irrelevant to the translation of a specific word.  
    \end{compactitem}
    
    
    Typically, the global noises will bring adverse effects for the whole translation, leading to incorporating noisy words, e.g. words ``just'' and ``in'' in Table~\ref{table-motivation}
    While the global noises are usually easy to notice, the local noises are tricky and receive less attention.
    We notice that, even when there is no global noise, some external words may still affect negatively at a certain time step, resulting in wrong translations.
    E.g. in Table~\ref{table-motivation}, when generating the word ``moving'' in the example sentence, the external word ``workers'' (correct translation of \begin{CJK}{UTF8}{gbsn}``工人"\end{CJK}) is the local noise. As a result, handling of local noises is also essential.

    In this paper, we propose a general framework to tackle the noise problem for diverse scenarios of external information. Our framework employs two separate word discriminators for the two kinds of noises, respectively, i.e. a \textit{global word discriminator} and a \textit{local word discriminator}. The global discriminator decides whether the provided words are useful or not, and the local discriminator decides whether the words should be applied at the current translation step. Our framework is trained with synthetic training data generated by directly sampling words from parallel sentences, which requires no additional data or manual annotation. 
    Experiments are conducted on two language pairs, two neural architectures, and four real-world scenarios where the external information could be machine translation results, lexical table of an SMT system, word-based translation from a bilingual dictionary or simply bag-of-words of the target sentence.
    
    
    We get the following conclusions:
    \begin{compactitem}
        \item The noises indeed prevents NMT models from benefiting more from external information. 
        \item Discriminating the noises leads that our model significantly outperforms the one without discrimination in translation quality as a consequence of better incorporating the external information, especially in very noisy conditions.
        \item Once the model is trained on the synthetic dataset, it can be directly used to improve different real-world scenarios without any task-specific tuning. It also indicates that the form of external words generalizes well to cover various types of external information.
    \end{compactitem}

\section{Related Work}
     Previous work focuses on integrating a certain kind of external information. For example, 
interactive machine translation systems could now employ assistance from humans, which could be as simple as one single correction of the translation~\cite{knowles2016neural,peris2017interactive,Hokamp:2017:Lexically}. 
 Some studies try to integrate external dictionaries into the NMT models~\cite{Luong:2014:Addressing, Arthur:2016:Incorporating, Li:2016:Towards} or improve the translation based on extra parallel data or the output of other translation systems ~\cite{zhou-EtAl:2017:Short1,niehues2016pre,Gu2017}.
    Unlike these work, we study a more general form of external information, which is applicable to different scenarios. 

    Besides, most of the previous methods require the presence of specific resources for training, e.g. translation of the parallel data generated by existing MT system(s)~\cite{zhou-EtAl:2017:Short1,niehues2016pre}. \newcite{wang2017neural,wang-EtAl:2017:EMNLP20173} propose approaches to use an SMT model to provide word and phrase recommendations for an attention-based NMT, where the two systems are deeply coupled. \newcite{wang2017exploiting,tu2018cache,Voita2018Context} propose to train context-aware translation models by the aids of large document/discourse-level data. In contrast, our training procedure is more general and simpler, which only uses word sampling from the original parallel data and requires no external resources.

    

To leverage outside information, such as words, for a generation task,
\newcite{Hokamp:2017:Lexically, Post2018Fast, hasler2018neural} propose to use lexical constraints on decoding process to utilize correct external word translations. \newcite{P16-1154} propose to use copying mechanism in the single-turn dialogue task, inspiring us for the basic framework. Compared to their attempts, our approach provides more robust solutions to discriminate noises. 
     


\section{Notation and Background}
    \subsection{Notation}
        We use the following notations throughout this paper. We denote a source sentence as $X =\langle x_1, \dots, x_{I}\rangle$, and a target sentence as $Y=\langle y_1, \dots, y_{T}\rangle$. The external words is denoted as  $Y^{E} = \{ y^{E}_1, \dots, y^{E}_{J}\}$. Because we focus on the case where $Y^{E}$ is a set of words, no sequential relation between words in $Y^{E}$ is considered, which reduce the requirement for external information. 
        This assumption makes the proposed methods applicable to wider applications, where the external words may be arbitrary.

\subsection{Neural Machine Translation}
    Traditional NMT systems use an \textit{encoder-decoder} architecture with \textit{attention mechanism} for translation ~\cite{bahdanau2014neural,D15-1166}. 
    The specific neural structure could be Recurrent Neural Network (RNN)~\cite{Bahdanau:2017:ICLR}, Convolutional Neural Network (CNN)~\cite{Gehring2017ConSeq} or the self-attention network (Transformer) ~\cite{Vaswani2017Attention}. 
    
    NMT models the translation probabilities from the source sentence to the target sentence in a word-by-word manner: 
    \begin{equation}
        P(Y|X) = \prod_{t=1}^{T} P(y_t|y_{<t},\mathbf{x})
    \end{equation}
    
    In first, the encoder maps the source sentence $\mathbf{x}$ into distributed representations $\bm{H} = \langle \bm{h}_1, \dots, \bm{h}_I\rangle$.
    In the $t$-th step of the decoding process, the word translation is generated by the decoder according to the following probability:
    \begin{equation}\label{eq:nmt}
        P(y_t|y_{<t},X) =\mathrm{softmax}(g(y_{t-1}, \bm{s}_t, \bm{c}_t))
    \end{equation}
    where $g(\cdot)$ is a non-linear activation function; $y_{t-1}$ is the output word of time step $t-1$; $\bm{s}_t$ is the current decoder hidden state which is modeled as: 
    \begin{align}
        \bm{s}_t=f(y_{t-1}, \bm{s}_{<t}, \bm{c}_t) 
        \label{eq:dec_state}
    \end{align}
    where $f(\cdot)$ is a transforming function depending on certain architectures; $\bm{c}_t$ is the source context vector from the attention mechanism:
    \begin{align}
        \bm{c}_t &= \sum_{i=1}^{I}{\alpha_{t,i} \cdot \bm{h}_i}
        \label{eqn-context} \\
        \alpha_{t,i} &= \mathrm{softmax}\big(a(\bm{s}_{t}, \bm{h}_i)\big)
        \label{eqn-alignment-probability} 
    \end{align}
    where $a(\cdot)$ is the attention model for the relation between $\bm{s}_{t}$ and the $i$-th source representation $\bm{h}_i$.

\section{A Basic Reading-Fusion Framework}
\label{sec:framework}
    
        
    The encoder-decoder architecture generates translation in a word-by-word manner. As a result, the incorporation of external words also affects the translation word-by-word, i.e. at each time step. Here we first present a basic and simple \textit{reading-fusion} framework, inspired by the structure of the Pointer Networks~\cite{vinyals2015pointer,P16-1014} and the Copying Mechanism~\cite{P16-1154}. 

    \paragraph{Reading stage} At each decoding time step $t$, an attention is performed between the concatenation of current decoding state $\bm{s}_t$ and source context $\bm{c}_t$, and the embedding of each external words.
    \begin{align}
        q_{t,j}^E &= \mathrm{softmax}\big(a([\bm{s}_{t}; \bm{c}_{t}], \bm{E}(y^{E}_j)) \big)
        \label{eq:att-probs-external}
    \end{align}
    The resulting attention weight $q^E_{t,\cdot}$ is treated as a probability distribution over all external words at time step $t$. That is, $P^E(y_j|X) = q_{t,j}^E$. The higher $P^E(y_j|X)$ is, the more related $y^E_j$ is to the current translation.
    
    \paragraph{Fusion stage} Similar to \newcite{Gu2017}, the probability distribution $P^E(y_j|X)$ is then interpolated with the original word generation probability from the decoder to perform the integrated word generation:
    \begin{align}
        \!\!\!\!\!\! P(y_t|X,\! Y^{\!E}) \!\! = \!\! (1 \!\! -  \!\! \beta_t) P(y_t|y_{<t},X) \! + \! \beta_t  P^E(y_t|X) \!\!
        \label{eq:newprob}
    \end{align}
    The scalar fusion gate $\beta_t$ is used to determine the relevance between the external content $\bm{c}^E_t$ and the translation at the current time step. $\beta_t$ is computed based on the representation of external content (external context vectors $\bm{c}^{E}_t$) and the representations of current step (decoding state $\bm{s}_t$ and source context vectors $\bm{c}_t$):
    \begin{align}
        \beta_t &= f_{\beta}(\bm{s}_t; \bm{c}_t; \bm{c}^E_t)
        \label{eq:gating} 
    \end{align}
    where $f_{\beta}$ is feed-forward neural networks with sigmod activation. The external context vector $\bm{c}^{E}_t = \sum_{j=1}^{J}{q_{t,j}^E \cdot \bm{E}(y^{E}_j)}$ is a weighted sum of the embeddings of the external words.

\begin{figure}[t] 
    \centering
    \includegraphics[width=0.5\textwidth]{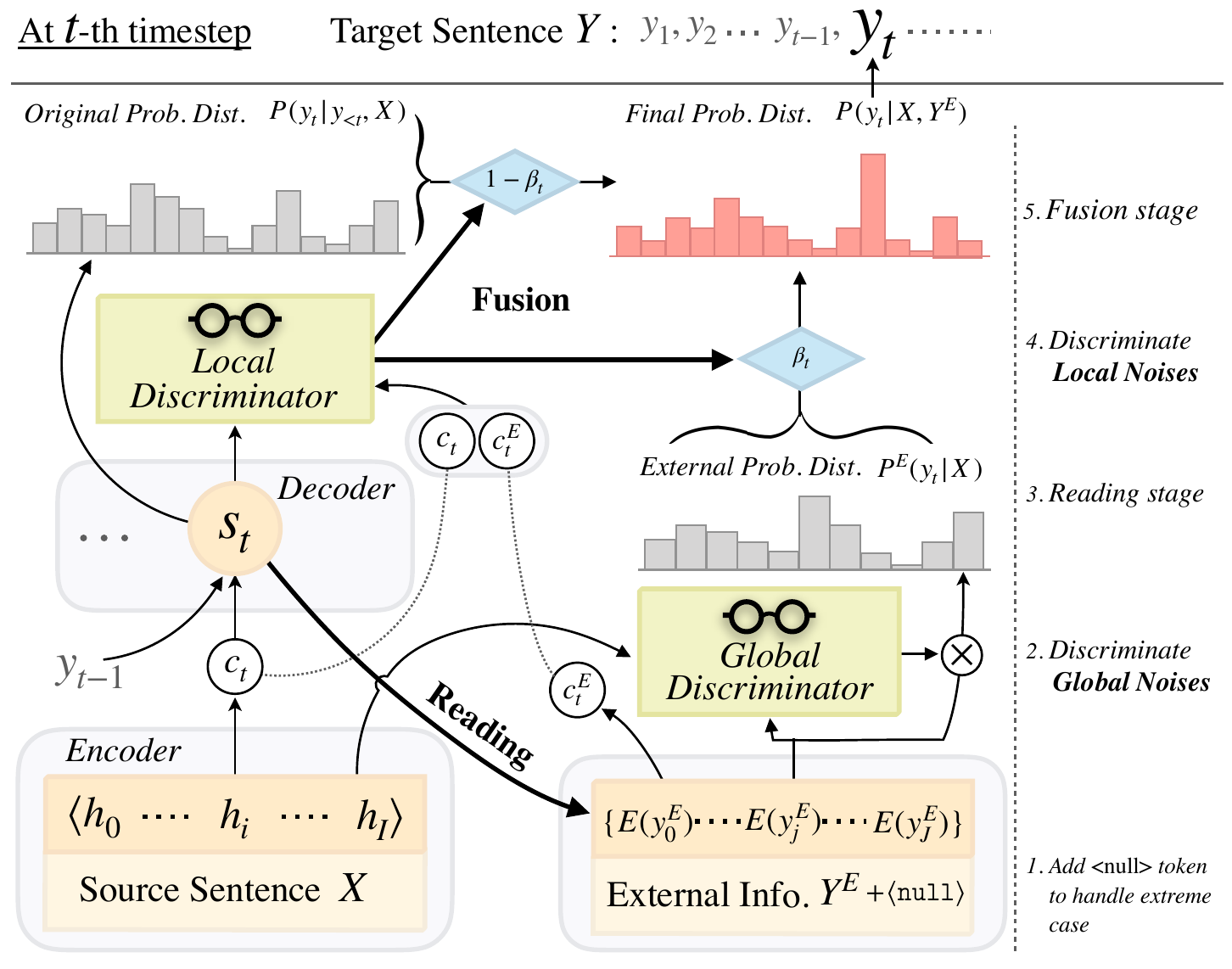}
    \caption{The Reading-Fusion framework with Noises Discrimination.}
    \label{fig:model2}
\end{figure}

\section{Discriminating the Noise in External Information}
    The reading-fusion framework could be interpreted as a way to copy the external word $y_j^E$ as the next generated word, with the probability determined by $P_{t,\cdot}^E$, and $\beta_t$. 
    However, when the external words are noisy, the integration process could be affected. 
    In this section, we describe specific approaches to discriminate two different kinds of noise, i.e. global noise and local noise.
    
    Figure~\ref{fig:model2} illustrates the proposed architecture. 
    Before decoding, a supervised global word discriminator is designed to determine whether each given external words is relevant to the current translation~(Section \ref{sec:global}). During decoding, in the reading stage, an attention mechanism is performed to select the correct external words or an extra \texttt{<null>} token; in the fusion stage, a supervised local word discriminator decides whether to use the obtained information for the translation of current word~(Section \ref{sec:local}).

    \subsection{Discriminating the Global Noise} 
        Because the source sentence and the external words are given before translation, a natural way of preventing the influence of the global noise is identifying them before translation.
        Here we propose a global word discriminator for filtering noisy input at the sentence level.
            
        \subsubsection*{Global Word Discriminator}
        \label{sec:global}
            Given a source sentence $X$, for each external word $y_j^E$, the global word discriminator makes the decision based on the embedding of current external word $\bm{E}(y^{E}_j)$, the source summarization $\bm{z}= \sum_{\bm{h_i}\in\bm{H}} \bm{h}_i / |\bm{H}| $ from the encoder~\cite{Sennrich2016}, and the attentive context vector $\bm{c}^D_{j}$, computed by an attention between the external word $y^E_j$ and source hidden states $\bm{H}$:
            \begin{align}
            \label{eq:gwd}
                D(y^{E}_j) &= f_{global}(\bm{E}(y^{E}_j); \bm{z} ; \bm{c}^D_{j}) 
            \end{align}
            where $f_{global}$ is feed-forward neural networks with sigmod activation. 
            \paragraph{Integration} We integrate the results of the global word discriminator into the NMT translation by discounting the word generation probability $q^E_{t,j}$ of the external word $y^{E}_j$ by $D(y^{E}_j)$. We revise the Equation~\ref{eq:att-probs-external} as follow:
            \begin{align}\label{eq:soft}
                \!\!\!q_{t,j}^E &= \mathrm{softmax}\big( a([\bm{s}_{t}; \bm{c}_{t}], \bm{E}(y^{E}_j)) \cdot D(y^{E}_j) \big) \!
            \end{align}
            As a result, the external words determined as global noises by the discriminator will have lower probabilities.
            


        \paragraph{Learning}
            Instead of training the decision of global word discriminator as a hidden variable with the whole model, we provide direct supervision to ensure its effectiveness.
            %
            %
            With the training instances, the global word discriminator is trained by minimizing the following cross entropy loss:
            \begin{align}
                loss^{g} =& \sum_{j=1}^{J}\big(-b(y^E_j)\!\cdot\! \log(D(y^{E}_j))  \label{eq:global-loss} \\
                        &\quad-\!(1\!-\!b(y^E_j)) \!\cdot\!\log(1\!-\!D(y^{E}_j)) \big)   \nonumber \\
                \mathrm{where} &\quad b(y^{E}_j) =
                \begin{cases}
                    \; 0\quad y^E_j \notin Y \nonumber \\
                    \; 1\quad y^E_j \in Y
                \end{cases} 
            \end{align}
            $b(y^{E}_j)$ indicates if the given external word $y^{E}_j$ is in the reference $Y$.

    \subsection{Discriminating the Local Noise}
        
        
        Even without global noise, the external words irrelevant to the current decoding time step may still attract attention mistakenly. To prevent the decoder from this unexpected influence, we propose to use a supervised local word discriminator to discriminate these noises. Additionally, let's imagine an extreme case, where there is no relevant external word in the current time step. All external words should be considered to be the local noises in the case, for which we propose an extra \texttt{<null>} token to distract the attention. 
        
        
        
        \subsubsection*{Local Word Discriminator}
        \label{sec:local}
            In the basic framework, the fusion gate $\beta_t$ automatically learns to distinguish the relevant information from the irrelevant one for a certain time step. 
            To encourage better decisions, we add a discriminative supervision for $\beta_t$. We can now refer the fusion gate to local word discriminator. 
        
            At each time step $t$ during training, the word choice $y_t$ is determined by the reference, which we use as the supervised information. In cases where the external words ${Y}^{E}$ contain $y_t$, the external context is considered relevant. Thus the local word discriminator is trained to predict positive. Otherwise, the external context is surely irrelevant, where the local word discriminator is trained to predict negative.
            
            
            
            
            \paragraph{Learning}
                The local word discriminator is trained via the following cross entropy loss:
                \begin{align}
                    loss^l =&\sum_{t=1}^{T}\big(-b(y_t) \cdot \log(\beta_t)  \label{eq:local-loss}\\
                        &\qquad- (1 - b(y_t)) \cdot \log(1-\beta_t) \big) \nonumber \\
                    \mathrm{where} &\quad b(y_t)=
                    \begin{cases}
                        \; 0& y_t \notin Y^E \nonumber \\
                        \; 1& y_t \in Y^E
                    \end{cases} 
                \end{align}
                $b(y_t)$ indicates if $y_t$ is in the external words $Y^{E}$. 
            

        
        \paragraph{Handling the extreme case}
        \label{sec:null}
            When the set of external words does not contain correct translation for the current target word, a natural idea is that \textit{no} word is helpful at all. Therefore, we add a special \texttt{<null>} token, which is expected to draw higher attention when there is no proper word available, into the set of external words. We do not use the probability $p^E(\texttt{<null>}|X)$ to generate target word. The token decreases the weights of the irrelevant words, easing the burden of local word discriminator.


    \subsection{Model Training}
        Given the dataset of training triples $\{\langle X_{m}, Y_{m}, Y^E_{m}\rangle \}_{m=1}^{M}$, the model parameters are trained by minimizing the loss $\mathcal{L}(\theta, \theta_g, \theta_l)$, where $\theta, \theta_g, \theta_l$ are the parameter of the NMT, global discriminator and local discriminator, respectively.
    	\begin{align}
    	\label{eq:new-obj}
    		\mathcal{L}(\theta, \theta_g, \theta_l)\!=\!\frac{1}{M}\!\!\sum_{m=1}^{M}\!\big(&-\!\log P_{\theta}(Y_{m}|X_{m}, Y^E_{m}) \nonumber \\
    		&+\lambda_1 \cdot loss^{g}_{\theta_g} \nonumber \\
    		&+\lambda_2 \cdot loss^{l}_{\theta_l} \big)
    	\end{align}
    	where $\lambda_1$ and $\lambda_2$ are hyper-parameters.

\begin{algorithm}
    \footnotesize
    \DontPrintSemicolon
    \caption{Construction of the synthetic dataset}  
    \label{alg:data}
    \KwIn{Parallel dataset $\mathcal{D}_1 = \{\langle X_{m}, Y_{m}\rangle\}_{m=1}^{M}$, vocabulary of target language $V$}
        \KwInit{$\mathcal{D}_2 = \emptyset$} \;
        \ForEach{sentence pair $\langle X, Y\rangle \sim \mathcal{D}_1$}{
            $Y^E = \emptyset$\;
            Randomly sample a value of p-ratio $\zeta \sim \mathrm{U}(0, 1)$\;
            Randomly sample $ceil(\zeta * |Y|)$ positive words from $Y$, and add them to $Y^E$\;
            Randomly sample $ceil((1-\zeta) * |Y|)$ negative words from $V \setminus Y$, and add them to $Y^E$\;
            Add a \texttt{<null>} token to $Y^E$ \;
            Update dataset: $\mathcal{D}_2 \leftarrow \mathcal{D}_2 \cup
            \langle X, Y, Y^E \rangle$\;
        }
        \KwRet $\mathcal{D}_2$\;
\end{algorithm}
	
    \section{Constructing Synthetic Training Dataset by Sampling}
        
        In order to train our model with proposed discriminating components, we propose a self-generated approach to construct a synthetic dataset of external words. Given parallel corpus $\mathcal{D}_1 = \{\langle X_{m}, Y_{m}\rangle\}_{m=1}^{M}$, synthetic external words are constructed for each sentence as training data, by sampling words in the reference as positive words, and the rest words in the vocabulary as negative words, respectively. Therefore, no additional data or annotation is required for training. Here we measure the volume of external words by the ratio of the number of provided words to the length of the sentence, denoted as \textit{v-ratio}:
        \begin{equation}
            v\text{-}ratio = \frac{|Y^E|}{|Y|}=\frac{\#\mathrm{posWord} + \#\mathrm{negWord}}{|Y|} \nonumber
        \end{equation}
        We measure the quality of the external words by the ratio of positive words to the total provided words, denoted as \textit{p-ratio}:
        \begin{equation}
            p\text{-}ratio \!= \!\! \frac{\#\mathrm{posWord}}{|Y^E|} \!\!=\!\! \frac{\#\mathrm{posWord}}{\#\mathrm{posWord} \!+\! \#\mathrm{negWord}}  \nonumber
        \end{equation}
        All models are trained using synthetic external words with v-ratio 1.0 and uniformly sampled p-ratio. See Alg. \ref{alg:data} for more details.

\begin{table*}[ht] 
\centering
\begin{tabular}{clcccccc} 

\#   & {\bf Model}                                  & MT03       & MT04      & MT05      & MT06      & Tests Avg.      & $\Delta$\\ 
\toprule 
0   &   \textsc{RNNSearch}                          &    32.00    &   33.92   &   30.08   &   28.21   &   31.48   &   -	\\
1   &   \textsc{Reading-Fusion} [\textsc{Basic}]                              &    33.49 &	34.67 &	32.09 &	30.04 &	32.27 &	+1.53\\
\hline
2   &   +\textsc{Global}~~[\textsc{Oracle}]          &    35.60 &    37.50 & 34.12 & 31.90 & 34.51 & +3.77\\
3   &   +\textsc{Global}                             &    35.59 &	37.00 &	33.57 &	31.57 &	34.05 &	+3.31\\
\hline
4   &   +\textsc{Local}                              &    35.24 &	36.26 &	32.99 &	31.90 &	33.72 &	+2.98  \\
5   &   +\texttt{<null>}                             &    34.68 &	35.25 &	32.93 &	31.74 &	33.31 &	+2.57\\
6   &   +\textsc{Local} \& \texttt{<null>}             &    35.65 &	37.00 &	33.69 &	32.00 &	34.05 &	+3.49\\
\hline
7   &   +\textsc{Global\&Local}\&\texttt{<null>}~[\textsc{Final}]     &    \textbf{36.50}	 &	 \textbf{37.45}	&	\textbf{34.34}	&	\textbf{32.50}	&	\textbf{34.76}	&  \textbf{+4.03}\\
\hline
\end{tabular} 
\caption{Comparison of different components of our approach (the last column presents the relative improvement comparing to the \textsc{RNNSearch} baseline).}
\label{tab:zh-en_bleu} 
\end{table*}

\section{Experiment}
We conduct experiments on Chinese-to-English (Zh-En) and English-to-German (En-De) translation tasks, respectively. 
For Zh-En, the training data consists of 1.6 million sentence pairs extracted from LDC\footnote{The corpora includes LDC2002E18, LDC2003E07, LDC2003E14, Hansards portion of LDC2004T07, LDC2004T08 and LDC2005T06}. We use NIST \texttt{MT03} dataset as the development set; \texttt{MT04}, \texttt{MT05}, \texttt{MT06} datasets as the test sets. The Chinese part of the data is segmented into words using ICTCLAS\footnote{http://ictclas.nlpir.org/}.
For En-De, we use WMT17~\cite{WMT:2017} corpus, which consists of 5.6M sentence pairs. We use \texttt{newstest2016} as our development set, and \texttt{newstest2017} as our testset.
We follow \newcite{edinWMT17:arxiv} to segment both German and English words into subwords using byte-pair encoding~\cite[BPE]{Sennrich2016Neural}. We use the merged vocabulary after BPE for both languages.

The translation evaluation metric is case-insensitive BLEU \cite{papineni2002bleu} for Zh-En\footnote{\url{https://github.com/moses-smt/mosesdecoder/blob/master/scripts/generic/multi-bleu.perl}}, and case-sensitive BLEU for En-De\footnote{\url{https://github.com/EdinburghNLP/nematus/blob/master/data/multi-bleu-detok.perl}}, which are consistent with previous work.
To evaluate and analyze the proposed approaches, we perform two categories of experiments on both synthetic and real-world settings. We first conduct experiments on the synthetic testsets similar to the training set. The aim of synthetic experiments is for analysis and ablation studies, where the experimental conditions could be easily manipulated. Then, we perform four real-world settings to evaluate the robustness and generality of our approach in practice. Note that in real-world settings, we use the same trained models from our main experiments without task-specific tuning for the different given datasets.

    \paragraph{Training details}
        For Zh-En, we limit the vocabulary size to 30K words, while we keep full vocabularies for En-De.
        All the out-of-vocabulary words are mapped to a special token \texttt{<\small UNK>}. 
        The value of $\lambda_1$ and $\lambda_2$ in Equation~\ref{eq:new-obj} are empirically set to 0.1, respectively.
        We train each model with sentences no longer than 50 words. The word embedding dimension is 512 and the size of all hidden layers is 1024. 
        The training batch size is 80. The beam size is set to 5 for testing.
        Training are performed via Adadelta~\cite{Zeiler2012ADADELTA} on a single GTX1080. 
        
            
    \subsection{Model Comparison and Analysis}
        \subsubsection*{Ablation Study on Zh-En Translation}
        For the ablation study, we compare different components of our model and list the results in Table~\ref{tab:zh-en_bleu}.
        Table~\ref{tab:zh-en_bleu} shows that when external words are available, all the models can make use of the assistance with different abilities.
        \paragraph{Discriminating noises are essential} We can see that ignoring noises inside the external words only leads to a moderate improvement (line 1). Discriminating either global or local noises improves the baseline by 3.31 (line 3) and 3.49 (line 6) BLEU scores, respectively. Particularly, with the learned global discriminator, our model achieves close performance compared to the oracle model (line 2 v.s. line 3), which can be regarded as the up-bound of discriminating the global noise with the ground-truth decisions.
        Our final model (line 7) combines all the proposed components to handle both global and local noises simultaneously. It obtains the highest 4.03 BLEU improvement. The final model gains stronger performance than those with single component, which indicates that handling global and local noises are both essential, and complement to each other.

        \begin{table}[t]
            \centering
            \begin{tabular}{lcc}
                {\bf Model}		                                & newstest2016 & newstest2017     \\
                \toprule
                \textsc{RNNSearch}                              &   29.3 &   23.5 \\
                \hline
                \textsc{Basic}                                 &   29.8  & 24.0 \\
                \textsc{Final}        &   {\bf 31.4} & {\bf 25.2}\\
            \hline
            \end{tabular}
            \caption{BLEU scores on En-De translation.}
            \label{tab:de-en}
        \end{table}

    \subsubsection*{Different Language Pair and Architecture} 
    We further validate the generality of the proposed model that we apply additional experiments on En-De translation for cross-language generality, and on neural Transformer-based NMT~\cite{Vaswani2017Attention} for cross-architecture generality.
    
        \paragraph{WMT17 En-De translation}
        Table~\ref{tab:de-en} shows the same trend as Zh-En translation task. This result indicates the effectiveness and generality of our method across various language pairs and translation granularities (words and BPE units).
            
        \paragraph{Transformer-based architecture}
        We also extend our approach to the recent emerging sequence-to-sequence architecture, \textsc{Transformer} model, on Zh-En task. Table \ref{tab:transformer} shows the consistent improvement with experiment on \textsc{RNNSearch}, which demonstrates the proposed approach is transparent to neural architectures, leading to feasible extension to other sequence-to-sequence models, such as and CNN-based model~\cite{Gehring2017ConSeq}.
            
    \begin{table}[t]
            \centering
            \begin{tabular}{lcc}
                {\bf Model}		                                & Dev. (MT03) & Tests Avg.     \\
                \toprule
                \textsc{Transformer}                             &   41.48 &   39.53 \\
                \hline
                \textsc{Basic}                                 &   42.91  & 40.78 \\
                \textsc{Final}        &   {\bf 45.52} & {\bf 42.34}\\
                \hline
            \end{tabular}
            \caption{BLEU scores with Transformer-based NMT.}
            \label{tab:transformer}
        \end{table}

\begin{figure}[t]

\centering
\subfloat[Fixed p-ratio 0.5 with different v-ratios.]{
\label{fig:volume}
\includegraphics[width=0.35\textwidth]{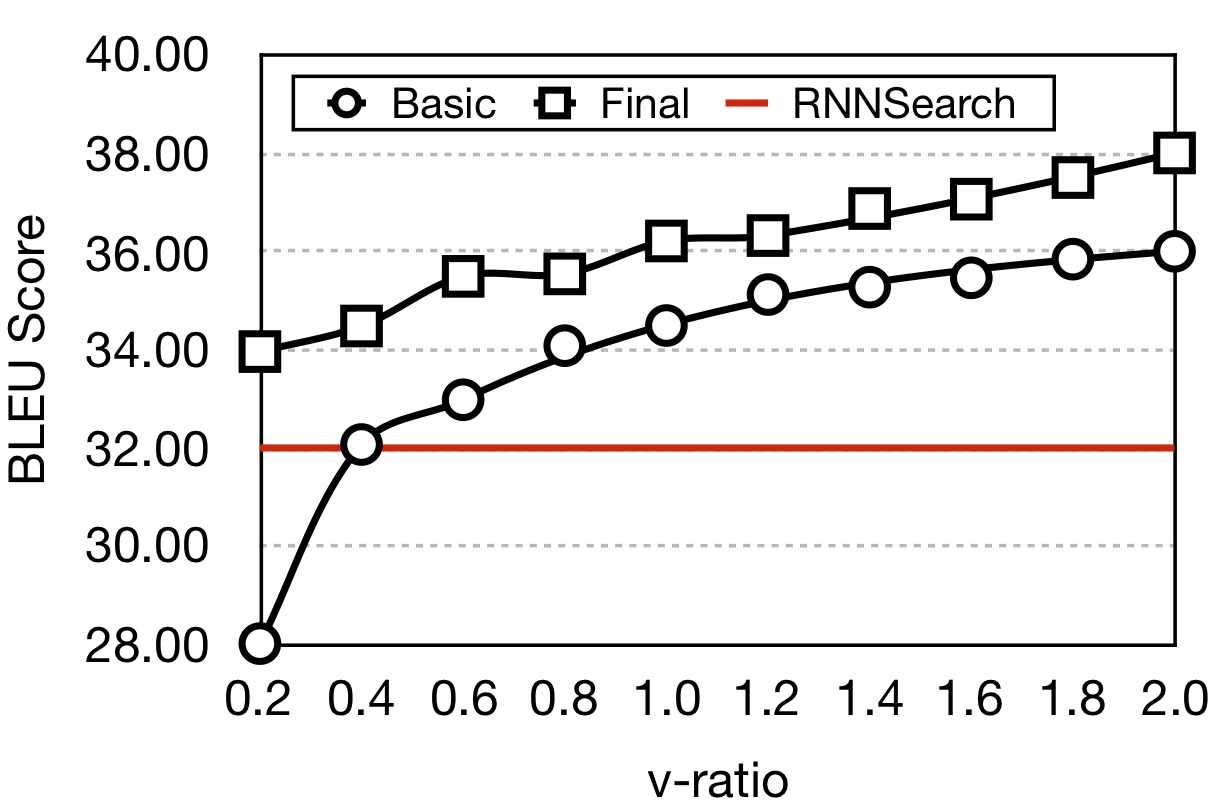}}\\

\subfloat[Fixed v-ratio 1.0 with different p-ratios.]{
\label{fig:rate}
\includegraphics[width=0.35\textwidth]{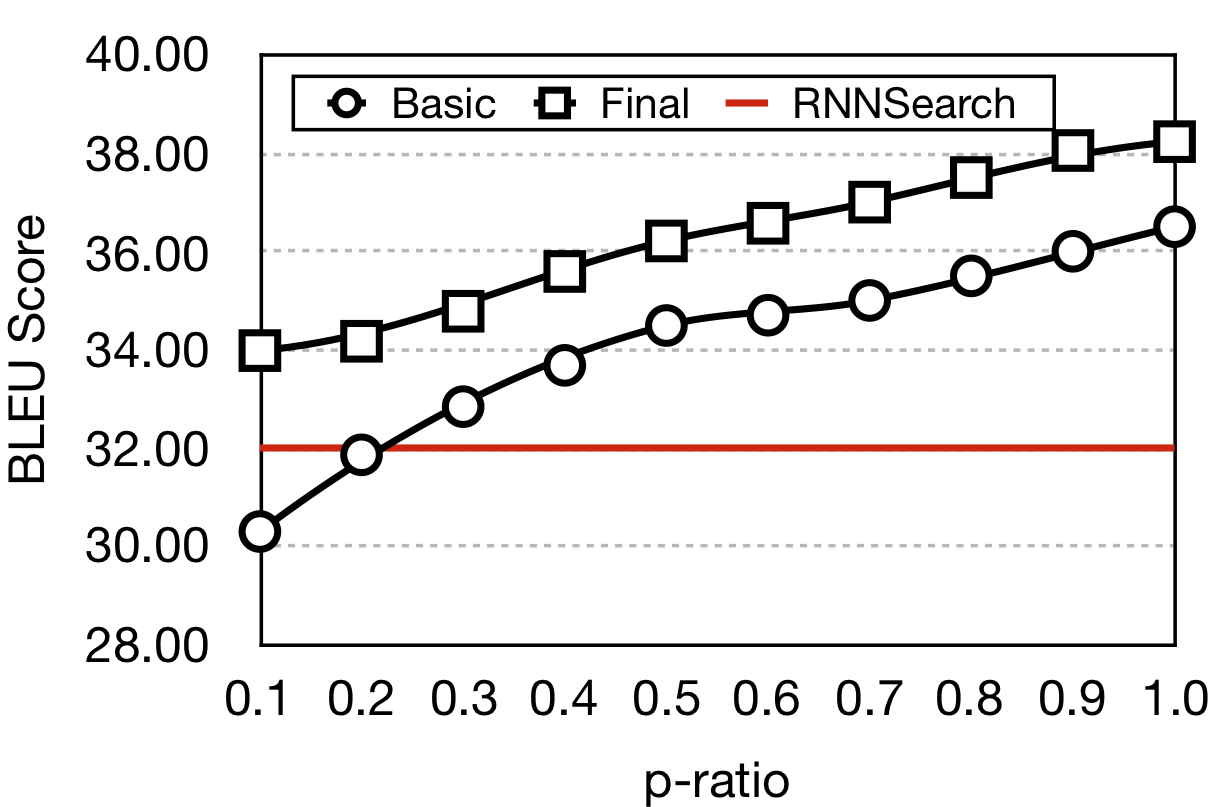}}

\caption{BLEU scores with varied ratio of external words.}

\end{figure}

    \subsubsection*{Study on Varied Ratio of External Words}\label{ss:ratio}
        To test the performance of our trained model under different conditions, we evaluate the translation quality improvement with different volume (v-ratio) and quality (p-ratio) of external words, respectively. For each experiment, we fix one of the ratio parameters and varies the other one. 
        As shown in Figure \ref{fig:volume} and \ref{fig:rate}, when the provided information is very short (v-ratio 0.2) or noisy (p-ratio 0.1), the \textsc{Basic} methods, only using simple attention and gating which is not able to process the noisy information, may not be able to improve the baseline model, showing the double-sided effect of noisy external words. However, our \textsc{Final} model could successfully discriminate the useful words from the noises and bring stable improvement in all the experimented conditions.

    \begin{table}[t]
        \centering
        \begin{tabular}{m{50pt}cc}
            
            {\bf }		        & Dev. (MT03) & Tests Avg.\\
            \toprule
            {\bf Precision} & 91.98\% & 92.33\% \\
            {\bf Recall} & 86.07\% & 86.37\% \\
            {\bf F-1 score} & 88.92\% & 89.25\% \\
            \hline
        \end{tabular}
        \caption{Performance of the global word discriminator.}
        \label{tab:discr-acc}
    \end{table}

    \subsubsection*{Performance of the Global Word Discriminator}
    
    To evaluate the prediction performance of the global word discriminator, we keep a held-out set of instances during training. Table \ref{tab:discr-acc} shows the precision, recall and F-1 score of the global word discriminator on this held-out set.
    The results show that on all test sets, the global discriminator achieves an F-1 score higher than 89\%, showing that the discriminator is indeed able to distinguish noises from useful external words. 

        \begin{table}[t]
            \centering
            \begin{tabular}{lcccc}
                {\bf Model}		                               & \textsc{SMT}         & \textsc{Lex}  & \textsc{Dict} & \textsc{BoW} \\
                \toprule
                \textsc{RNNSearch}                             &   \multicolumn{4}{c}{32.00} \\
                \hline
                \textsc{External}                              &   30.67      &   8.24 &   6.85  &   - \\
                \hline
                \textsc{Basic}                                 &   33.28      &   32.57 & 32.23 & 32.98 \\
                \textsc{Final}        &   {\bf 35.71} & {\bf 34.10} & {\bf 33.23} &   {\bf 34.34} \\
            \hline
            \end{tabular}
            \caption{BLEU scores comparison on real-world scenarios.}
            \label{tab:smt}
        \end{table}

    \subsection{Experiments with Real-World Scenarios}
        We also test our system in four real-world scenarios, where four different sources of information are used as external inputs. 
        \paragraph{Statistic machine translation (SMT)} 
        We firstly use the translations results from the Moses SMT system~\cite{koehn2007moses} as external words. The SMT translation result is often considered to be complete because it translates the whole sentence with less under-translation problems, while is also complained to be less fluent and contain more errors, where de-noising is quite important.
        
        \paragraph{Lexical table (\textsc{Lex})} 
        Besides directly using the translation outputs from Moses, the intermediate products such as the lexical table could also be leveraged to conduct word-level translations where each source word is mapped separately to its translation.
        
        \paragraph{Bilingual dictionary (\textsc{Dict})} Bilingual dictionary is relatively easy to obtain, compared with other resources such as other translation systems, human interaction or parallel corpora. We investigate the case where word-level translations are directly used as external information. The most severe type of noises of both \textsc{Lex} and \textsc{Dict} is that they are literally word-level translations without morphological changes in the context of the current sentence.
        
        \paragraph{Bag-of-Words prediction (\textsc{BoW})}
        \newcite{Weng:2017:EMNLP} propose a multi-tasking scheme to boost NMT by predicting the bag-of-words of target sentence using the proposed Word Predictions approach. We are curious if the predicted bag-of-words have the potential to help the NMT. Here we follow the WPE configuration in their paper to train a word predictor, where the target bag-of-words are predicted by the encoder's summarization $\bm{z}$ (see \newcite{Weng:2017:EMNLP} for details). The word predictor is trained on the top of our models, whose parameters are all frozen. We collect the top-K predicted words from its prediction where $K=1.0 \times |X|$ for each source sentence. 
        
        Note that, as same as Section~\ref{ss:ratio}, we directly use the trained model without any specific training on the given datasets. Interestingly and surprisingly, Table \ref{tab:smt} shows the importance of de-noising in all the four scenarios, in which our model derives greater benefits from the noisy real-world external datasets. Furthermore, we present two practical examples in Table \ref{tab:case}. These evaluations give a further demonstration of the effectiveness and generality of our discriminating approaches.

    \begin{CJK}{UTF8}{gbsn}
    \begin{table}[t]
    \footnotesize
    \renewcommand\arraystretch{1.2}
    \centering
        \subfloat[Human correction with one word as external information.]{\begin{tabular}{m{32pt}|m{165pt}} 
            \toprule  
            \textsc{Source}            & 已有三对染色体完成排序，包括第二十对、第二十一对和第二十二 对。\\ 
            \textsc{Refer.}            & the sequencing of three chromosomes has been completed, including chromosomes 20, 21, and 22.\\
            \textsc{NMT}    & there have been three \textit{\texttt{<\small UNK>}, namely }, the 20th, and \it{ \texttt{<\small UNK>}, on \texttt{<\small UNK>}}.\\
            \hline
            \hline
            \textsc{Extern.}  & {\bf chromosome}   \\
            \hline
            \textsc{Final}    &  three {\bf chromosome} have been completed, including the 20, 21 and 22. \\
            \bottomrule
        \end{tabular}
        \label{tab:case-1-word}}\\
        \subfloat[SMT output as external information.]{\begin{tabular}{m{32pt}|m{165pt}} 
            \toprule  
            \textsc{Source}            & 我们不愿为以往达成的协定再度付出代价。 \\ 
            \textsc{Refer.}            & we are not willing to pay again for the agreements that have been reached already.\\
            \textsc{NMT}&  we do not want to pay the price for \it{a further price}.\\
            \hline
            \hline
            SMT                 & we do not want to pay the price for {\bf the agreements reached in the past}.   \\
            \hline 
            \textsc{Final}    &  we do not want to pay the price for {\bf the agreements we have reached.}\\
            \bottomrule
        \end{tabular}
        \label{tab:case-smt}}
    \caption{Case study on de-noising and incorporating two different sources of external information. Surprisingly, correct revision could further improve the translation followed by the previous errors.}
    \label{tab:case}
    \end{table}
\end{CJK}

\section{Conclusion and Future Work}

    
    In this paper, we focus on the noise problem when NMT models are able to access and incorporate external information. There are two kinds of noises in external information, i.e., the global and local noises. We propose a general framework that learns to discriminate the both noises directly on a self-generated synthetic dataset that requires nothing external but the original parallel data. We find that the noises indeed prevents NMT models from benefiting more from external information. In experiments, our noise discrimination shows its superiority in the incorporation of external information, especially in very noisy conditions. Further analysis indicates that our model can be directly used without any task-specific tuning in the various scenarios, where the patterns of noises are different. It also indicates that the form of external words generalizes well to cover various types of external information. For future work, it may be interesting to adapt our noise discrimination for sequentially encoded external information.


\bibliography{naaclhlt2019}
\bibliographystyle{acl_natbib}

\end{document}